\documentclass[conference,10pt]{IEEEtran}
\usepackage{amsmath}
\usepackage{amsfonts}
\usepackage{amssymb}
\usepackage{setspace}
\usepackage{graphicx}
\usepackage{amsthm}
\usepackage{color}
\usepackage{caption}
\usepackage{pgf}
\usepackage{subcaption}
\usepackage{cite}
\usepackage{epsfig,psfrag}
\usepackage{tabularx}
\usepackage{pgf}
\usepackage{tikz}
\usepackage{pst-node}
\usepackage[yyyymmdd,hhmmss]{datetime}
\usepackage{notation}
\usepackage{psfrag}
\usepackage{pstool}
\usetikzlibrary{arrows,backgrounds,calc,positioning,shapes,shadows}
\usepackage[T1]{fontenc}
\usepackage[utf8]{inputenc} 

\providecommand{\ist}{\hspace*{.3mm}}
\providecommand{\rmv}{\hspace*{-.3mm}}

\providecommand{\nn}{\nonumber}
\newcommand{\T}{\mathrm{T}}
\definecolor{temporalgreen}{RGB}{0,128,0}
\definecolor{spatialred}{RGB}{255,0,0}
\definecolor{temporalblue}{RGB}{0,0,205}

\allowdisplaybreaks

\begin{document}
\title{Moment-Matching Probabilistic Data Association\\  for Optimization-Based SLAM\vspace{-1mm}}

\author{
    \IEEEauthorblockN{
        Khoa Nguyen$^\ast$, 
        Mitchell Turton$^\ddagger$,
        and Florian~Meyer$^\ast$
    }
    \vspace*{.5mm}

    \IEEEauthorblockA{
        $^\ast$Department of Electrical and Computer Engineering, 
        University of California San Diego, USA\\
        $^\ddagger$Georgia Institute of Technology, Atlanta, GA\\
        {Email: ktnguyen@ucsd.edu,  mturton3@gatech.edu, and flmeyer@ucsd.edu.}
    }
\vspace{-7.5mm}
}

\maketitle
\begin{abstract}
Optimization-based simultaneous localization and mapping  (SLAM) makes it possible to reduce accumulated navigation errors of sensing platforms by returning to known areas (``loop closure''). In this paper, we present an approach to combine probabilistic data association (PDA) with optimization-based SLAM. Instead of associating a single measurement with each landmark, we follow the PDA paradigm from the multiobject tracking community. In particular, in a processing stage performed in addition to the nonlinear least-squares solver of optimization-based SLAM, our method (i) assigns multiple measurements to landmarks probabilistically, (ii) computes the mean and covariance of landmark distributions via moment matching by taking multiple measurement-to-landmark associations into account, and (iii) establishes a virtual landmark measurement and a corresponding linear-Gaussian measurement model that leads to the mean and covariance matrix as moment-matching PDA in (ii). By converting the PDA update step into an equivalent linear-Gaussian measurement update step, PDA can be performed effectively within any optimization-based SLAM method. Our preliminary numerical evaluation in a scenario with false negatives and false positives indicates that incremental smoothing and mapping 2 (iSAM2), combined with the proposed PDA approach, can improve agent localization performance compared to conventional iSAM2.

\end{abstract}
\begin{IEEEkeywords}
SLAM, data association, state estimation, localization, mapping, smoothing
\vspace{.5mm}
\end{IEEEkeywords}


\section{Introduction}\label{sec:intro}

For landmark-based navigation, simultaneous localization and mapping (SLAM) \cite{Dellaert2006SquareRootSAM,thrun2006graph,kaess2008isam,kaess2012isam2,dellaert2012factorGTSAM,CadCarCar:J16,BowAtaDanPap:C17,Hsiao2019MHiSAM2,EbaBerBig:J23,slam-handbook,LeiMeyHlaWitTufWin:J19,LeiVenTeaMey:J23,GeKalXia:J25,LiaLeiMey:J25} aims to jointly estimate an agent's trajectory and a map of the surrounding environment using sensor measurements. A key aspect of many SLAM applications is loop closure: when the agent returns to previously mapped landmarks, the estimation error for all agent states on the loop is significantly reduced. While loop closure is essential for reducing estimation errors, it requires the ability to update the joint state involving all landmark states and all agent states on the loop. This contrasts with multi-object tracking problems and related SLAM methods \cite{LeiMeyHlaWitTufWin:J19,LeiVenTeaMey:J23,GeKalXia:J25,LiaLeiMey:J25}, where typically only the most recent state of objects or landmarks is updated.  Optimization-based SLAM \cite{Dellaert2006SquareRootSAM,thrun2006graph,kaess2008isam,kaess2012isam2,dellaert2012factorGTSAM} can address the high-dimensional estimation problem with loop-closure constraints while maintaining favorable computational complexity by linearizing the system dynamics and solving the resulting linear-Gaussian estimation problem with least-squares solvers.

 Optimization-based SLAM methods, including batch SLAM and iSAM, rely on nonlinear least-squares optimization in which motion and observation measurements from sensors are incorporated as measurement factors in a factor graph \cite{Dellaert2006SquareRootSAM,thrun2006graph,kaess2008isam,kaess2012isam2,dellaert2012factorGTSAM}. Under Gaussian noise assumptions, these approaches provide accurate and computationally efficient state estimation and have become the standard for long-duration autonomy in agent deployments.

A fundamental challenge in SLAM is data association (DA), which determines which measurements correspond to previously observed landmarks. Traditional optimization-based SLAM typically performs ``hard'' DA using global nearest-neighbor assignments \cite{Dellaert2006SquareRootSAM,thrun2006graph,kaess2008isam,kaess2012isam2,dellaert2012factorGTSAM}. This approach is directly compatible with popular nonlinear least-squares solvers but can be unreliable in the presence of false negatives (FNs) and false positives (FPs). It also results in overconfident uncertainty quantification. Incorrect hard associations are difficult to recover from, resulting in potential track loss and degraded map consistency \cite{sh-ch3-outlier}. In particular, traditional optimization-based SLAM is reduced to nonlinear least squares where the squared-residual objective causes a small number of grossly incorrect measurements to dominate the quadratic cost and drive the optimizer toward an incorrect trajectory and map estimate \cite{sh-ch3-outlier}. 



Robustness techniques are commonly divided into front-end  \cite{FisBol:C81,ManDomEusVas:C18}  and back-end approaches \cite{SunPro:C12,AgaTipSpi:C13}. Front-end approaches serve as preprocessing steps that reject outliers in sensor data by enforcing consistency constraints. Prominent examples include Random Sample Consensus (RANSAC) \cite{FisBol:C81} and pairwise consistency maximization (PCM) \cite{ManDomEusVas:C18}. Back-end approaches are robust optimization methods that mitigate the influence of remaining bad constraints \cite{sh-ch3-outlier}. A widely used approach is to replace the squared loss with a robust loss to reduce, resulting from the Gaussian assumption of measurement noise, to reduce the contribution of large residuals by using losses with sub-quadratic growth \cite{PenKumVid:C23}. These robust objectives are commonly optimized using Iterative Reweighted Least Squares (IRLS) \cite{PenKumVid:C23}. In SLAM, this can improve resilience to outlier constraints. However, depending on the chosen kernel, the problem may become more nonconvex and can exhibit increased sensitivity to initialization and local minima \cite{sh-ch3-outlier}.   

PDA \cite{BarLi:95,BarWilTia:B11} addresses limitations related to hard associations by incorporating all possible measurement hypotheses within a Bayesian update step. This update step relies on a moment-matching approximation to transform the Gaussian mixture into a single Gaussian \cite{BarLi:95,BarWilTia:B11}, thereby correctly quantifying association uncertainty. However, PDA is not directly compatible with optimization-based SLAM methods. After marginalizing over all possible soft associations, the likelihood function of PDA consists of a weighted sum of individual likelihood functions, which cannot directly be used within a least-squares optimization framework. Classical JPDA explicitly enumerates all possible global associations. This scales exponentially with problem size and becomes infeasible in moderate-sized problems \cite{MeyKroWilLauHlaBraWin:J18}. Scalable alternatives rely on Markov Chain Monte Carlo (MCMC) techniques \cite{Del:T01} and loopy belief propagation (BP) \cite{MeyKroWilLauHlaBraWin:J18}. Soft associations have been previously considered for SLAM problems within an expectation maximization (EM) framework \cite{Dav:J07,BowAtaDanPap:C17,Michael2022PDASemanticSLAMScale}. EM is guaranteed to converge to a local maximum but cannot provide correct uncertainty quantification \cite{JamJen:J02}. While not derived in the EM framework, the method in \cite{GeKalXia:J25}, similar to EM, iterates between sampling a likely single hard association for all time steps and performing optimization-based SLAM. The multi-hypothesis extension of iSAM2 \cite{Hsiao2019MHiSAM2} aims to maintain and optimize multiple hypotheses over time steps, similarly to the multi-hypothesis tracker (MHT) \cite{Willett2007}.

\begin{figure*}
        \centering
        \includegraphics[scale=.65]{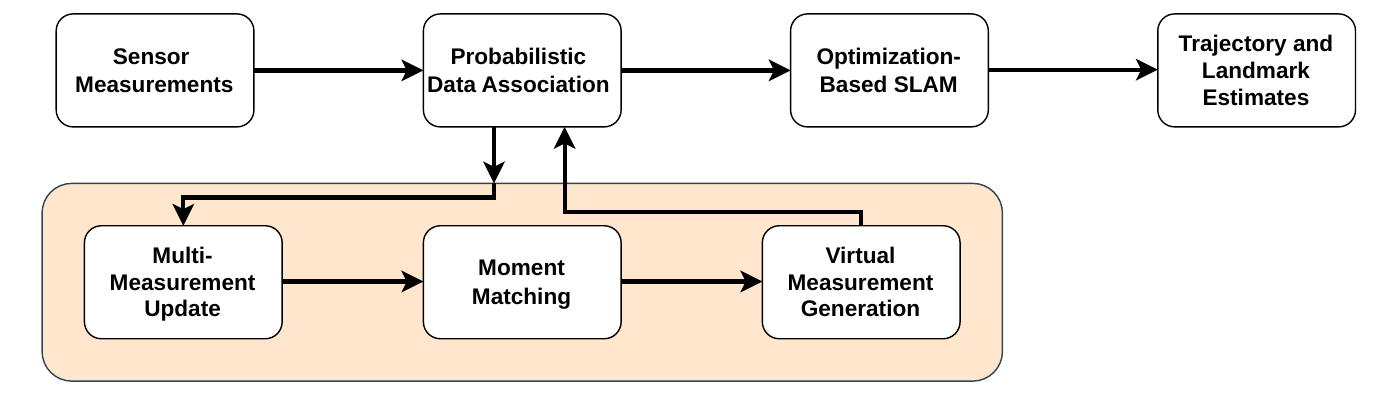}
                \caption{Flow diagram showing the proposed pipeline for integrating PDA with optimization-based SLAM.}
        \label{fig:block}
        \vspace{-2mm}
\end{figure*}

This paper presents a PDA approach for optimization-based SLAM that converts a PDA update step, based on all possible measurement hypotheses, into a virtual measurement and a corresponding linear-Gaussian measurement model.  The information contribution of the virtual measurement exactly matches that of the PDA update step. The resulting measurement allows the association uncertainty to be incorporated into optimization-based SLAM. Our approach can be interpreted as performing expectation propagation (EP) \cite{minka2005divergence} on the subgraph of the SLAM problem \cite{LeiMeyHlaWitTufWin:J19} that includes the variable nodes representing random associations.  Our preliminary numerical results indicate that the proposed method can improve localization and mapping performance in scenarios with a significant number of FPs and FNs\vspace{0mm}.

\section{Background on Optimization-Based SLAM}
\label{sec:background slam}

In optimization-based SLAM, the statistical model of the estimation problem is typically represented by a factor graph with motion constraints and measurements defined as probabilistic factors that connect state variables \cite{dellaert2012factorGTSAM,kaess2008isam,KscFreLoe:01}. A key feature of optimization-based SLAM is loop closure. Loop closure is essential for reducing estimation errors but it requires the ability to update the joint state that involves all landmark states and all agent states on the loop.

\subsection{Batch SLAM}
Batch Simultaneous Localization and Mapping (Batch SLAM) formulates localization and mapping as a global optimization problem in which the entire agent trajectory and all landmark states are estimated simultaneously \cite{thrun2006graph,Dellaert2006SquareRootSAM}. The Batch SLAM method incorporates all available measurements into a single least-squares system, allowing information from later observations to refine earlier agent state estimates.

Let the joint state vector be defined\vspace{1mm} as
\begin{equation}
\V{s} = \big[ \V{x}^{\T}_1, \V{x}^{\T}_2, \dots, \V{x}^{\T}_K, \V{\ell}^{\T}_1, \V{\ell}^{\T}_2, \dots, \V{\ell}^{\T}_J \big]^{\T}
\label{eq:state vector}
\vspace{1mm}
\end{equation}
where $\V{x}_k \in \mathbb{R}^d_x$ denotes the agent state at time step $k$,
for $k=1,\dots,K$, and $\V{\ell}_j \in \mathbb{R}^\ell$ is the state of landmark $j$ that represents the position of the landmark.
The trajectory of the agent consists of agent states $\V{x}_{1:K}$, and $\{\V{\ell}_j\}_{j=1}^{J}$ denotes the set of landmarks.  Assuming Gaussian noise, the maximum a posteriori (MAP) estimate is obtained by minimizing a weighted nonlinear least-squares objective
\begin{equation}
\V{s}^\star =
\arg\min_{\V{s}}\;
\| \V{x}_1 - \V{\mu}_1 \|_{\M{\Sigma}_1^{-1}}^2 +
\sum_{k=2}^{K}
\| \V{x}_k - g(\V{x}_{k-1},\V{u}_k) \|_{\M{\Sigma}_{\V{u}}^{-1}}^2 \nn
\end{equation}

\vspace{-2mm}
\begin{equation}
+\sum_{m=1}^{M}
\| \V{z}_m - h(\V{\ell}_{j_m} , \V{x}_{k_m}) \|_{\M{\Sigma}_{\V{z}}^{-1}}^2
\label{eq:batch_cost}
\vspace{.8mm}
\end{equation}
where $\V{\mu}_1$ and $\M{\Sigma}_1$ denote the prior mean and covariance of the initial agent state, $\V{z}_m$ denotes the $m$-th measurement observation associated with agent state $\V{x}_{k_m}$ and landmark $\V{\ell}_{j_m}$, and $M$ is the total number of measurements. Each term in \eqref{eq:batch_cost}, corresponds to a factor of the Gaussian factor graph representing the statistical model of the SLAM problem \cite{KscFreLoe:01}.

After linearization about the current estimate, the nonlinear model reduces to a linear least-squares problem \cite{Dellaert2006SquareRootSAM}
\begin{equation}
\V{s}^\star
=
\arg\min_{\V{s}}
\left\lVert \M{A}\V{s} - \V{b} \right\rVert^2
\label{eq:batch_linear_system}
\end{equation}
where $\M{A}$ is a sparse Jacobian matrix encoding all constraints and $\V{b}$ is the stacked measurement residual vector. The batch least-squares system is formed by stacking the linearized constraint matrices associated with the prior, motion, and observation factors \cite{dellaert2012factorGTSAM}. The resulting global Jacobian has the following\vspace{0mm} structure $\M{A} = [\M{P}^{\T}\;\M{G}^{\T}\;\M{H}^{\T}]^{\T}$. Here, $\M{P}$ represents the prior factor, $\M{G}$ encodes odometry-based motion constraints between consecutive agent states, and $\M{H}$ corresponds to landmark measurement constraints. Each constraint is weighted according to its measurement uncertainty.

The optimal state estimate $\V{s}^\star$ is obtained by solving the linear least-squares problem using either QR factorization or Cholesky\vspace{.5mm} decomposition \cite{Dellaert2006SquareRootSAM}
\begin{equation}
\M{A}^\T \rmv \M{A}\V{s}^\star = \M{A}^\T \V{b}.
\label{eq: bslam state estimate}
\end{equation}QR factorization provides improved numerical stability, while Cholesky factorization exploits the symmetric positive-definite structure of the normal equations for computational efficiency. Both approaches yield identical MAP estimates when numerical conditioning is adequate \cite{Dellaert2006SquareRootSAM}. {A key observation is that the topology of the factor graph induces the sparsity structure of $\M{A}$. The fact that $\M{A}$ has significant sparsity leads to a computation complexity that is reasonable even for SLAM problems with thousands of random variables.

Although batch SLAM produces globally consistent estimates \cite{Dellaert2006SquareRootSAM}, its computational cost grows with deployment length because the entire nonlinear least squares problem must be solved again if a factor, e.g., resulting from a new measurement, is added. This limitation motivates the need for incremental approaches, which can update the track of the agent efficiently without solving a complete batch SLAM problem.

\subsection{Incremental Smoothing and Mapping (iSAM)}
\label{sec:isam}

iSAM methods aim to combine the ability to update the entire track of an agent with sequential processing as performed by a conventional filter. Both iSAM1 and iSAM2 can be developed based on QR factorization \cite{kaess2008isam,kaess2012isam2} of the Jacobian $\M{A}$, i.e.\vspace{-.5mm},
\begin{equation}
\M{A} =\M{Q}
\begin{bmatrix}
\M{R} \\
\M{0}
\end{bmatrix}.
\vspace{.5mm}
\label{eq:QR_equation}
\end{equation}
In (\ref{eq:QR_equation}), the matrix $\M{R}$ is an upper triangular matrix known as the square-root information matrix, which satisfies
\begin{equation}
\M{R}^\T \M{R} = \M{A}^\T \M{A}.
\end{equation}
Some of the sparsity structure of $\M{A}$ is preserved in $\M{R}$. The sparsity of $\M{R}$ is determined by the factor-graph topology and the chosen variable ordering. Good orderings reduce fill-in and are central to the computational efficiency of square-root SLAM methods. In iSAM1 and iSAM2, the square-root information matrix is updated incrementally by appending new measurement constraints to the system and performing local operations to restore the upper triangular structure \cite{kaess2008isam,kaess2012isam2}. 

In iSAM1 \cite{kaess2008isam} the upper-triangular structure is restored using the Givens rotations. Givens rotations modify only the affected portions of $\M{R}$. In pure exploration problems with local measurements, the number of Givens rotations per update can remain bounded, leading to approximately constant-time updates. However, loop closures can introduce fill-in in $\M{R}$, making later updates more expensive. For this reason, iSAM1 periodically performs a periodic batch refactorization/reordering step in which the accumulated measurement Jacobian is rebuilt and refactorized using batch SLAM. iSAM1 periodically performs a batch reordering and refactorization step. The variable ordering is chosen to reduce fill-in in the triangular factor $\M{R}$, and nonlinear measurement functions can be relinearized during these batch steps. The new, sparse square-root information matrix $\M{R}$ is then used as the starting point for subsequent incremental updates  \cite{kaess2008isam}. 

iSAM2 \cite{kaess2012isam2} can avoid any periodic batch refactorization/reordering steps by incrementally reordering variables and selective relinearization of only the variables affected by new measurements. The key concept that enables the iSAM2 method is a graphical model called Bayes tree. The Bayes tree encodes the square-root factorization as a tree of conditional densities. When new factors are added, iSAM2 identifies the affected cliques, removes the affected subtree, combines the new and relinearized factors with orphaned subtrees, and re-eliminates this local portion of the graph. This enables incremental variable reordering and fluid relinearization without periodic batch refactorization.


\section{PDA for Optimization-Based SLAM}
\label{sec:PDA}

\subsection{Review of PDA for SLAM}
\label{PDAReview}

\indent PDA provides a Bayesian framework for state estimation in the presence of DA uncertainty  \cite{BarLi:95,BarWilTia:B11}. The PDA approach has been adopted for SLAM within the framework of belief propagation (BP) \cite{LeiMeyHlaWitTufWin:J19,LeiVenTeaMey:J23,GeKalXia:J25,LiaLeiMey:J25}, but not for optimization-based SLAM with loop-closure capabilities \cite{Dellaert2006SquareRootSAM,thrun2006graph,kaess2008isam,kaess2012isam2,dellaert2012factorGTSAM}. We consider a scenario where multiple measurements may be available at each time step, and it is unknown which measurement, if any, corresponds to a given landmark. PDA accounts for this uncertainty by probabilistically weighting all candidate measurements rather than choosing a single hard association. 

Let us assume that at time $k$, Gaussian representations of predicted posterior PDFs $f( \V{x}_k, \V{\ell}_j \ist | \ist   \V{z}_{1:k-1}) =  \mathcal{N}( \overline{\V{\mu}}_{k,j},\overline{\M{\Sigma}}_{k,j}) $, $j = 1,\dots,J$ are available. The covariance matrix $\M{\Sigma}_{k,j}$ can be computed from the estimate and square-root information matrix $\M{R}$ provided by optimization-based SLAM using dynamic programming \cite{kaess2008isam,kaess2012isam2}. In addition, a sensor returns a vector of measurements $\V{z}_k = \big[ \V{z}^{\T}_{k,1}, \dots, \V{z}^{\T}_{k,m_k} \big]^{\rmv\T}\rmv\rmv\rmv$. Each measurement either originates from a mapped landmark or is a FP. In particular, for landmark $j$ and under hypothesis $a^{(j)}_k = m$, a landmark-generated measurement is modeled as
\begin{equation}
\V{z}_{k,m} = \V{h}(\V{x}_k,\V{\ell}_{j}) + \V{n}_{k,m},
\qquad
\V{n}_{k,m} \sim \mathcal{N}(\V{0},\M{\Sigma}_{\V{z}}).
\label{eq:measModel}
\end{equation}

Each landmark is detected with probability $p_{\mathrm{d}}$, and the number of FP measurements follows a Poisson PMF with mean $\mu_{\mathrm{fp}}$. FP measurements are assumed independent and identically distributed according to $f_{\mathrm{fp}}(\V{z}_{k,m})$ \cite{BarWilTia:B11}.

Because the origin of each measurement is unknown, an association variable is introduced for each landmark. For landmark $j$ at time step $k$, the discrete association variable is defined\vspace{.5mm} as
\begin{equation}
a_k^{(j)} =
\begin{cases}
m, & \text{if meas. } z_{k,m} \text{ is generated by landmark } j,\\
0, & \text{if landmark } j \text{ is not detected}.
\end{cases} \nn
\end{equation}
Since it is unknown which measurement corresponds to which landmark, $a^{(j)}_k$ is a random variable.

An accurate BP-based approximation \cite{ LeiMeyHlaWitTufWin:J19} of the joint posterior PDF $f(\V{x}_k,\V{\ell}_j \ist | \ist \V{z}_{1:k})$ can now be obtained\vspace{1mm} as 
\begin{align}
f(\V{x}_k,\V{\ell}_j \ist | \ist \V{z}_{1:k}) &\approx C(\V{z}_k) \ist f(\V{x}_k, \V{\ell}_j \ist | \ist  \V{z}_{1:k-1}) \nn\\[1.5mm]
&\hspace{5mm}\times \rmv\sum_{a_k^{(j)}=0}^{m_k} \rmv\rmv
g( \V{x}_k,\V{\ell}_j, a_k^{(j)};\V{z}_k) \ist \nn
\kappa_k\big(a_k^{(j)}\big) \\[-4mm]
 \label{eq:pda_update} \\[-7mm]
 \nn
\end{align}
where the function
\begin{equation}
g( \V{x}_k,\V{\ell}_j, a_k^{(j)};\V{z}_k) 
=
\begin{cases}
\dfrac{p_{\mathrm d}\,
f(\V{z}_{k,m} \ist|\ist \V{x}_k,\V{\ell}_j)}
{\mu_{\mathrm{fp}} f_{\mathrm{fp}}(\V{z}_{k,m})},
& \hspace{-2.5mm} a_k^{(j)} = m, \\[3mm]
(1-p_{\mathrm d}),
& \hspace{-1.3mm} a_k^{(j)} = 0,
\end{cases}
\label{eq:gFunction}
\end{equation}
represents the measurement model, $C(\V{z}_k) $ is a constant that only depends on observed measurement $\V{z}_k$, and  $\kappa_k(a_k^{(j)})$ is a joint-association function that represents the influence of other objects in the environment.

\captionsetup[subfigure]{justification=centering, singlelinecheck=false, labelformat=empty}

\begin{figure*}[!t]
    \centering
    \begin{subfigure}[b]{0.32\textwidth}
        \centering
        \captionsetup{skip=2pt}
        \includegraphics[width=1\linewidth]{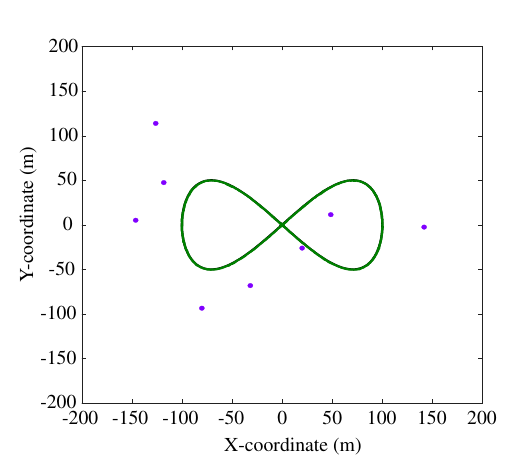}
        \subcaption{\hspace{6mm}(a)}
        \label{fig:ground truth}
    \end{subfigure} \hfill
    \begin{subfigure}[b]{0.32\textwidth}
        \centering
        \captionsetup{skip=2pt}
        \includegraphics[width=1\linewidth]{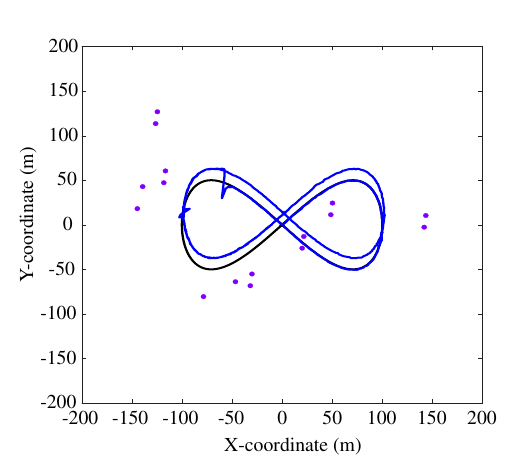}
        \subcaption{\hspace{6mm}(b)}
        \label{fig:data associaton result}
    \end{subfigure} \hfill
    \begin{subfigure}[b]{0.32\textwidth}
        \centering
        \captionsetup{skip=2pt}
        \includegraphics[width=1\linewidth]{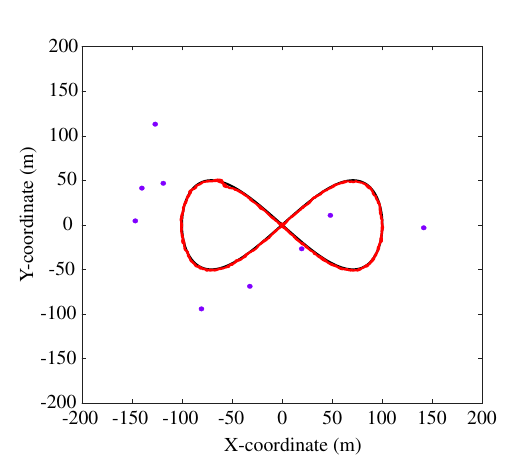}
        \subcaption{\hspace{6mm}(c)}
        \label{fig:pda result}
    \end{subfigure} \hfill
    \caption{Comparison of agent state estimation results using different association strategies. The agent follows a repeated figure-eight trajectory. Ground truth is shown in black. Estimated trajectories are shown in green for oracle batch SLAM, blue for DA + iSAM2, and red for PDA + iSAM2. Landmark estimates are shown as dots with uncertainty ellipses.}
    \label{fig:three graphs}
    \vspace{-4mm}
\end{figure*}

The term $(1-p_{\mathrm d})$ in \eqref{eq:gFunction} corresponds to the FN hypothesis $a_k^{(j)} \rmv = \rmv 0$, where landmark $j$ is not observed at time step $k$. When $a_k^{(j)}=m$, the likelihood term based on $f(z_{k,m} \ist | \ist \V{x}_k,\V{\ell}_j)$ favors agent and landmark states that align with measurement $z_{k,m}$. In  \eqref{eq:pda_update}, hypotheses are combined through a weighted sum over all measurements, so measurements that are more consistent with the predicted landmark observation model contribute more strongly \cite{BarLi:95,BarWilTia:B11}. The factor $\kappa_k(a_k^{(j)})$ reduces the weight of some association hypotheses when several landmarks ``compete'' for the same measurement \cite{MeyBraWilHla:J17,MeyKroWilLauHlaBraWin:J18,MeyWil:J21,JanMeySnyWigBauHil:J23}. It can be computed by brute-force marginalization \cite{BarLi:95}, MCMC techniques \cite{Del:T01}, or loopy BP \cite{MeyKroWilLauHlaBraWin:J18}. Note that we can interpret the expression $\sum_{a_k^{(j)}=0}^{m_k}
g( \V{x}_k,\V{\ell}_j, a_k^{(j)};\V{z}_k) \ist \nn
\kappa_k\big(a_k^{(j)}\big)$ on the right-hand-side of  \eqref{eq:pda_update} as a soft-association likelihood function.

Following the joint probabilistic data association (JPDA) paradigm \cite{BarLi:95,BarWilTia:B11}, a Gaussian representation $\mathcal{N}(\V{\mu}_{k,j},\M{\Sigma}_{k,j})$  of $f(\V{x}_k,\V{\ell}_j \ist | \ist \V{z}_{1:k})$ can be obtained by moment matching. After linearization, the PDA posterior is represented as a Gaussian mixture with $m+1$ components: one Kalman-updated component for each candidate measurement and one FN component corresponding to $a^{(j)}_k = 0$, for which the predicted Gaussian is retained. The mixture is next reduced to a single Gaussian $\mathcal{N}(\V{\mu}_{k,j},\M{\Sigma}_{k,j})$ with the same mean and covariance matrix by using well-known JPDA update equations \cite{BarLi:95,BarWilTia:B11}. This type of moment-matching approximation is referred to as EP in the graphical models literature \cite{minka2005divergence} .

A PDA update, as in  \eqref{eq:pda_update}, is well known to provide robustness in scenarios with FPs and FNs. However, it is not directly amenable to optimization-based SLAM, which expects a likelihood function that can be used within a least-squares objective \eqref{eq:batch_cost}. In the next section, we overcome this challenge by introducing a virtual landmark measurement and a corresponding linear-Gaussian measurement model that leads to the exact same mean $\V{\mu}_{k}$ and covariance matrix $\M{\Sigma}_{k}$ of $f(\V{x}_k,\V{\ell}_j \ist | \ist \V{z}_{1:k})$ as obtained by the soft association and moment matching approach discussed above.

\subsection{Proposed Virtual Measurement Generation}

Consider the predicted and updated posterior related to the PDA update step discussed in the previous section, i.e.,
\begin{align}
f(\V{x}_k,\V{\ell}_j \ist | \ist \V{z}_{1:k-1}) &= \mathcal{N}(\overline{\V{\mu}}_{k,j},\overline{\M{\Sigma}}_{k,j}), \nn\\[.7mm]
f(\V{x}_k,\V{\ell}_j \ist | \ist \V{z}_{1:k}) &= \mathcal{N}(\V{\mu}_{k,j},\M{\Sigma}_{k,j}).
\end{align}
with $\overline{\M{\Sigma}}_{k,j}  \succeq \M{\Sigma}_{k,j}  \succ 0$.

Our goal is to construct a virtual measurement model that is linear-Gaussian, i.e.,
\begin{equation}
  \V{y}_{k,j} = \M{H}_{k,j} \hspace{.1mm} [\V{x}^{\T}_k,\V{\ell}^{\T}_j]^{\T}  + \V{v}_{k,j}, \qquad
  \V{v}_{k,j} \sim \mathcal{N}(\V{0}, \M{I}),
\end{equation}
such that $f(\V{x}_k,\V{\ell}_j \ist | \ist \V{z}_{1:k-1},\V{y}_{k,j}) = f(\V{x}_k,\V{\ell}_j \ist | \ist \V{z}_{1:k})$, i.e., based on virtual measurement $ \V{y}_{k,j}$, we can reproduce the same Gaussian parameters $\V{\mu}_{k,j}$, $\M{\Sigma}_{k,j}$ obtained from the PDA update step as discussed in Section \ref{PDAReview} within a least-squares cost function as in \eqref{eq:batch_cost}. To simplify the following derivation, we restrict the additive noise $\V{v}_{k,j}$ to be zero-mean and with identity-covariance. We also drop the time index $k$ and the landmark index $j$ to simplify the notation. 

For a linear-Gaussian virtual measurement to reproduce the updated Gaussian posterior, the parameters of the predicted and posterior PDF must satisfy the Kalman update in information form for the considered noise model\vspace{0mm} \cite{ShaKirLi:B02}, i.e.,
\begin{align}
  \M{\Sigma}^{-1} &= \overline{\M{\Sigma}}^{-1} + \M{H}^\T \M{H}, \nn\\[1mm]
  \M{\Sigma}^{-1}\V{\mu} &= \overline{\M{\Sigma}}^{-1} \ist \overline{\V{\mu}} + \M{H}^\T \V{y}. \label{eq:virtualConstrains}
\end{align}
Next, we introduce the information gain $\M{J} = \M{\Sigma}^{-1} - \overline{\M{\Sigma}}^{-1}$ and information vector increment $\V{c} = \M{\Sigma}^{-1}\V{\mu} - \overline{\M{\Sigma}}^{-1}\ist \overline{\V{\mu}}$. It can easily be verified that $\M{J}$ is positive semidefinite with the same rank as $\overline{\M{\Sigma}}_k - \M{\Sigma}_k$. 

Based on \eqref{eq:virtualConstrains}, constructing the virtual measurement model reduces to finding $\M{H}$ and $\V{y}$ such that
\begin{equation}
  \M{J} = \M{H}^\T \M{H}, 
  \qquad
  \V{c} = \M{H}^\T \V{y}.
  \label{eq:infoForm}
\end{equation}
Since the information gain $\M{J}$ is positive semidefinite it admits an eigenvalue decomposition, i.e., $\M{J} = \M{U}\M{\Lambda}\M{U}^\T$, where $\M{\Lambda} = \mathrm{diag}(\lambda_1,\ldots,\lambda_d)$ with $\lambda_i \ge 0$. Let $r$ be the rank of $\M{J}$, and $\M{U}_r$ contain the eigenvectors
corresponding to eigenvalues $\lambda_i > 0$, i.e., $\M{J} = \M{U}_r \M{\Lambda}_r \M{U}_r^\T$ where $\M{\Lambda}_r \in \mathbb{R}^{r \times r}$ is diagonal and positive definite. Based on the square root of $\M{\Lambda}_r$, i.e., $\M{L}_r \rmv=\rmv \M{\Lambda}_r^{1/2}$, we can express the information gain as $ \M{J} = (\M{L}_r \M{U}_r^{\T})^{\rmv\T} (\M{L}_r \M{U}_r^\T)$. By comparing this expression with \eqref{eq:infoForm}, we directly obtain $\M{H} = \M{L}_r \M{U}_r^\T \in \mathbb{R}^{r \times d}$. Next, we use this expression for $\M{H}$ in the second equality in \eqref{eq:infoForm} and solve for $\V{y}$. In this way, we obtain the virtual measurement $\V{y} = \M{L}_r^{-1} \M{U}_r^\T \V{c}.$ 

It has been verified that if the consistency condition $\overline{\M{\Sigma}}  \succeq \M{\Sigma}  \succ 0$ is satisfied, we can use the virtual measurement $\V{y}$ and corresponding model $\M{H}$ in a least-squares cost function \eqref{eq:batch_cost} and obtain the PDA update $\V{\mu}$ and $\M{\Sigma}$ up to machine precision. Based on the PDA update step and the virtual measurement model, we can directly establish optimization-based SLAM methods that perform a PDA update step.

\section{Numerical Results}

In the simulations seen in Fig.~\ref{fig:three graphs}, the agent follows a repeated figure-eight track within a 400 by 400 meter environment that contains 10 randomly distributed landmarks. Landmark observations are limited to a sensing range of 100 meters from the agent and are modeled with additive zero-mean Gaussian noise with a standard deviation of 0.3 m. Velocity measurements are modeled with zero-mean Gaussian noise with a standard deviation of 0.3 m/s. The agent is initially located at the center of the scene and has perfect knowledge of its state. 

In our numerical validation, we considered both a MATLAB implementation of iSAM1 as well as iSAM2 provided by the GTSAM library \cite{dellaert2012factorGTSAM}. Since both methods lead to identical estimation results in the considered linear Gaussian problem, we only report iSAM2 results. In addition to the proposed iSAM2 combined with moment-matching PDA (``PDA + iSAM2''), we also simulate a reference method that performs hard data association (``DA + iSAM2'') using a global nearest-neighbor approach \cite{DuffKos:J01}. 
\begin{figure} [t]
        \centering
        \vspace{-3mm}
        \includegraphics[scale=1]{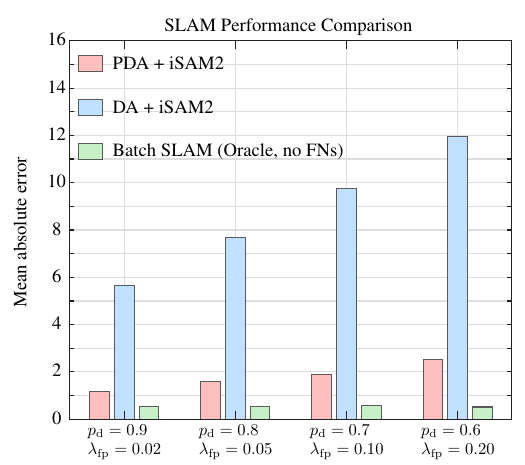}
        \caption{Comparison of mean absolute error for different association strategies for different FP and FN settings, parameterized by the probability of detection $p_\mathrm{d}$ and mean number of FPs $\mu_{\mathrm{fp}}$.}
        \label{fig:barplot}
        \vspace{-6mm}
\end{figure}
Here, hard assignment costs are computed as $-\log g\big( \V{x}_k,\V{\ell}_j, a_k^{(j)};\V{z}_k\big)$, $a_k^{(j)} \in \{0,\dots,m_k\}$. In this way, the parameters $p_\mathrm{d}$ and $\mu_{\mathrm{fp}}$ are correctly taken into account for hard assignments. It can easily be verified that for $p_{\mathrm d} = 1$ these hard assignment costs are equivalent to conventional costs based on the Mahalanobis distance. To evaluate performance in the presence of FPs and FNs, they are incorporated into the measurement process, where the probability of detection is set to $p_\mathrm{d} = 0.7$. False measurements are generated according to a Poisson process with a mean number of FPs of $0.1$.

The simulation uses a heuristic landmark management process to control landmark initialization. Landmarks are assigned using an m/n logic \cite{BarLi:95,BarWilTia:B11} with $m=3$ and $n=5$. Fig.~\ref{fig:three graphs} shows the true and estimated agent states for one simulation run.
The observed difference between Fig.~\ref{fig:pda result} when compared to Fig.~\ref{fig:data associaton result} demonstrates a significant improvement in agent state estimation when applying the PDA within the iSAM framework.

In Fig.~\ref{fig:barplot}, the mean absolute error is plotted and compared across three estimators: PDA + iSAM2, DA + iSAM2, and Batch SLAM (Oracle DA, no FNs). The oracle batch SLAM baseline is an idealized lower-error reference that uses ground-truth associations and all true landmark detections while excluding false positives. The comparison is performed for four $(p_\mathrm{d},\mu_\mathrm{fp})$ parameter settings, where $p_\mathrm{d}$ denotes the probability of detection and $\mu_\mathrm{fp}$ denotes the mean number of false alarms: $(0.9, 0.02)$, $(0.8, 0.05)$, $(0.7, 0.10)$, and  $(0.6, 0.20)$. These parameter changes produce increasingly more difficult DA conditions due to more FNs and higher levels of FP measurements. 300 simulation runs were performed.

Across all four cases, Batch SLAM with known measurement associations and no FNs yields the lowest mean absolute error and remains essentially unchanged across all parameter settings as its associations are not affected by FNs and FPs. When measurement associations are known, the method serves as the lowest error reference for comparison with the other approaches. Under low association uncertainty in Fig.~\ref{fig:barplot}, the PDA + iSAM2 and DA + iSAM2 methods are comparable to the Batch SLAM reference method in mean absolute error. However, as $p_\mathrm d$ decreases and $\mu_\mathrm{fp}$ increases, the difference in performance between the two DA methods changes significantly. The DA + iSAM2 method becomes less accurate under more difficult measurement conditions, resulting in higher mean absolute errors. On the other hand, the PDA + iSAM2 method shows improved robustness and better accuracy than the DA + iSAM2 method when the parameter conditions become more challenging. In the most challenging setting of $(p_\mathrm d,\mu_\mathrm{fp})=(0.6,0.20)$, PDA + iSAM2 achieves a lower mean absolute error of  2.515~m than DA + iSAM2 (11.961~m), while the batch SLAM reference remains at 0.513~m.  Table~\ref{tab:mae_results} provides the mean absolute error results for PDA + iSAM2, DA + iSAM2, and batch SLAM methods (Oracle DA, no FNs) under each $(p_\mathrm d,\mu_\mathrm{fp})$ setting\vspace{1mm}. 

\begin{table}[t]
\centering
\caption{Mean absolute error (in meters) for different $(p_\mathrm d,\mu_\mathrm{fp})$ parameter settings.}
\label{tab:mae_results}
\renewcommand{\arraystretch}{1.17}
\setlength{\tabcolsep}{3.6pt}
\footnotesize
\begin{tabular}{c c c c}
\hline
\textbf{$(p_\mathrm d,\mu_\mathrm{fp})$} & \textbf{PDA + iSAM2} & \textbf{DA + iSAM2} & \textbf{Batch (Oracle)} \\
\hline
$(0.9, 0.02)$ &  1.182 m
& 5.643 m
& 0.536 m
\\
$(0.8, 0.05)$ & 1.595 m
& 7.688 m
& 0.539m
\\
$(0.7, 0.10)$ & 1.887 m
& 9.761 m
& 0.572 m
\\
$(0.6, 0.20)$ &  2.515 m
& 11.961 m
&  0.513 m
\\
\hline
\end{tabular}
\vspace{-4mm}
\end{table}

\section{Conclusion and Future Work}
We proposed a probabilistic data association (PDA) approach for optimization-based simultaneous localization and mapping (SLAM) that (i) assigns multiple measurements to landmarks probabilistically, (ii) computes the mean and covariance of landmark distributions via moment matching by taking multiple measurement-to-landmark associations into account, and (iii) establishes a virtual landmark measurement and a corresponding linear-Gaussian measurement model that leads to the exact same mean and covariance as the PDA update.

The proposed method aims to provide robustness in scenarios with a significant number of false positives (FPs) and false negatives (FNs) where ``hard'' data association (DA) is challenged. Numerical results show that the proposed PDA + iSAM2 method exhibits improved robustness and lower mean absolute error than the reference DA + iSAM2 method as the probability of detection decreases and the mean number of false positives increases. In addition, our results show that the PDA + iSAM2 method maintains performance comparable to a batch SLAM reference, which performs perfect ``oracle-based'' DA with no FNs, in challenging measurement conditions. 

Future research will evaluate the proposed method on larger and more diverse simulations and extend the framework to more complex SLAM scenarios with nonlinear measurement models and three-dimensional agent motion. Additional work will also study computational tradeoffs and compare the proposed approach with other probabilistic and multi-hypothesis DA methods for SLAM. We will also focus on applying the framework to real-world datasets and deployments \cite{NguLiaDav:C25,DavNguJan:J25} and assess model robustness under realistic environmental and sensing uncertainty\vspace{-.5mm}.

\section*{Acknowledgment}
\vspace{.5mm}
This work was supported by the National Science Foundation (NSF) under CAREER Award No. 2146261. The authors thank Prof.~Frank Dellaert for illuminating discussions\vspace{-.5mm}.

\renewcommand{\baselinestretch}{1}
\footnotesize
\selectfont
\bibliographystyle{IEEEtran}
\bibliography{IEEEabrv,StringDefinitions,SALBooks,SALPapers,Temp}
\end{document}